\pdfoutput=1

\documentclass[11pt]{article}
\usepackage[table,xcdraw]{xcolor}

\usepackage[preprint]{acl}

\usepackage{times}
\usepackage{latexsym}

\usepackage[T1]{fontenc}

\usepackage[utf8]{inputenc}

\usepackage{microtype}

\usepackage{inconsolata}

\usepackage{graphicx}

\usepackage{amsmath}
\usepackage{amssymb}
\usepackage{pifont}
\definecolor{JSViolet}{RGB}{71,15,244}
\definecolor{JSRed}{RGB}{205,44,78}
\newcommand{\cmark}{\textcolor{JSViolet}{\ding{51}}}
\newcommand{\xmark}{\textcolor{JSRed}{\ding{55}}}
\usepackage[capitalize,noabbrev]{cleveref}
\usepackage{lipsum}
\usepackage{duckuments}
\usepackage{multirow}

\usepackage{booktabs}
\usepackage{subcaption}
\usepackage{xspace}
\usepackage{algorithm, algpseudocode}

\newcommand{\llname}{Mamba Drafters for Speculative Decoding}

\title{\llname}

\author{
  Daewon Choi$^{1}$,
  \,
  Seunghyuk Oh$^{1}$,
  \,
  Saket Dingliwal$^{2}$,
  \,
  Jihoon Tack$^{1}$,
  \,
  \\
  \textbf{Kyuyoung Kim}$^{1}$,
  \textbf{Woomin Song}$^{1, 2, \dagger}$,
  \textbf{Seojin Kim}$^{3, \ddagger}$,
  \textbf{Insu Han}$^{1}$,
  \textbf{Jinwoo Shin}$^{1}$,
  \\
  \textbf{Aram Galstyan}$^{2}$,
  \textbf{Shubham Katiyar}$^{2}$,
  \textbf{Sravan Babu Bodapati}$^{2}$
  \\
  $^{1}$KAIST
  $^{2}$Amazon AGI
  $^{3}$Seoul National University
}

\usepackage[symbol]{footmisc}

\begin{document}
\maketitle

\footnotetext{\ignorespaces$^{\dagger}$ Work done during an internship at Amazon. $^{\ddagger}$ Work done at KAIST.
}

\begin{abstract}
Speculative decoding has emerged as a promising approach to accelerating large language model (LLM) generation using a fast drafter while maintaining alignment with the target model's distribution.
However, existing approaches face a trade-off: external drafters offer flexibility but can suffer from slower drafting, while self-speculation methods use drafters tailored to the target model but require re-training.
In this paper, we introduce novel drafters based on Mamba, a state-of-the-art state space model (SSM), as a solution that combines the best aspects of both approaches.
By leveraging the linear structure of SSMs, our approach avoids the quadratic complexity inherent in traditional Transformer-based methods, enabling faster drafting and lower memory usage while maintaining the flexibility to work across different target models.
We further enhance efficiency with a novel test-time tree search algorithm for generating high-quality draft candidates.
Our empirical evaluation demonstrates that Mamba-based drafters not only outperform existing external drafting methods but are also comparable to state-of-the-art self-speculation approaches
while using less memory and maintaining their cross-model adaptability.
\end{abstract}

\section{Introduction}
\label{sec:intro}
Recent breakthroughs in large language models (LLMs) have been largely driven by Transformer architectures~\citep{vaswani2017attention}, which have enabled exceptional performance across a wide range of tasks~\citep{achiam2023gpt4,singhal2025medical,kim2024financial}.
However, their capabilities often come with significant computational overhead, primarily due to the autoregressive nature of sequential token generation, while the quadratic complexity of the attention mechanism further exacerbates scalability challenges. Speculative decoding (SD)~\citep{stern2018blockwise,leviathan2023fast,xia2023speculative,chen2023accelerating} has emerged as a promising approach to addressing the inefficiencies of autoregressive models by generating multiple candidate tokens with an efficient drafter and verifying them in parallel with the target model, ensuring identical output with greater efficiency.
This approach enables simultaneous decoding of multiple tokens within a single forward pass.

Existing SD methods are mainly categorized into two types: (i) using an external drafter \citep{leviathan2023fast} applicable to multiple target models, and (ii) adopting a self-speculation approach, where a drafter is trained to align with the target model itself, showing faster drafting speed compared to external drafters \citep{cai2024medusa,li2024eagle}. For instance, self-speculation involves training a small Transformer head on top of the target model’s large Transformer block to generate multiple candidate tokens, which are then used as a draft for the target model \citep{li2024eagle2}. While showing a faster drafting speed compared to using an external drafter, training a separate drafter for each target model is computationally expensive. More importantly, to handle distribution shift—i.e., to process novel inputs unseen during training—the drafter should be trained on a large corpus, which is particularly challenging because modifying only the last layer, a common practice, still requires forwarding all lower layers of the large target model during training, leading to significant computational overhead.

\begin{figure*}[t]
    \centering
    \begin{subfigure}{0.32\textwidth}
\includegraphics[width=\linewidth]{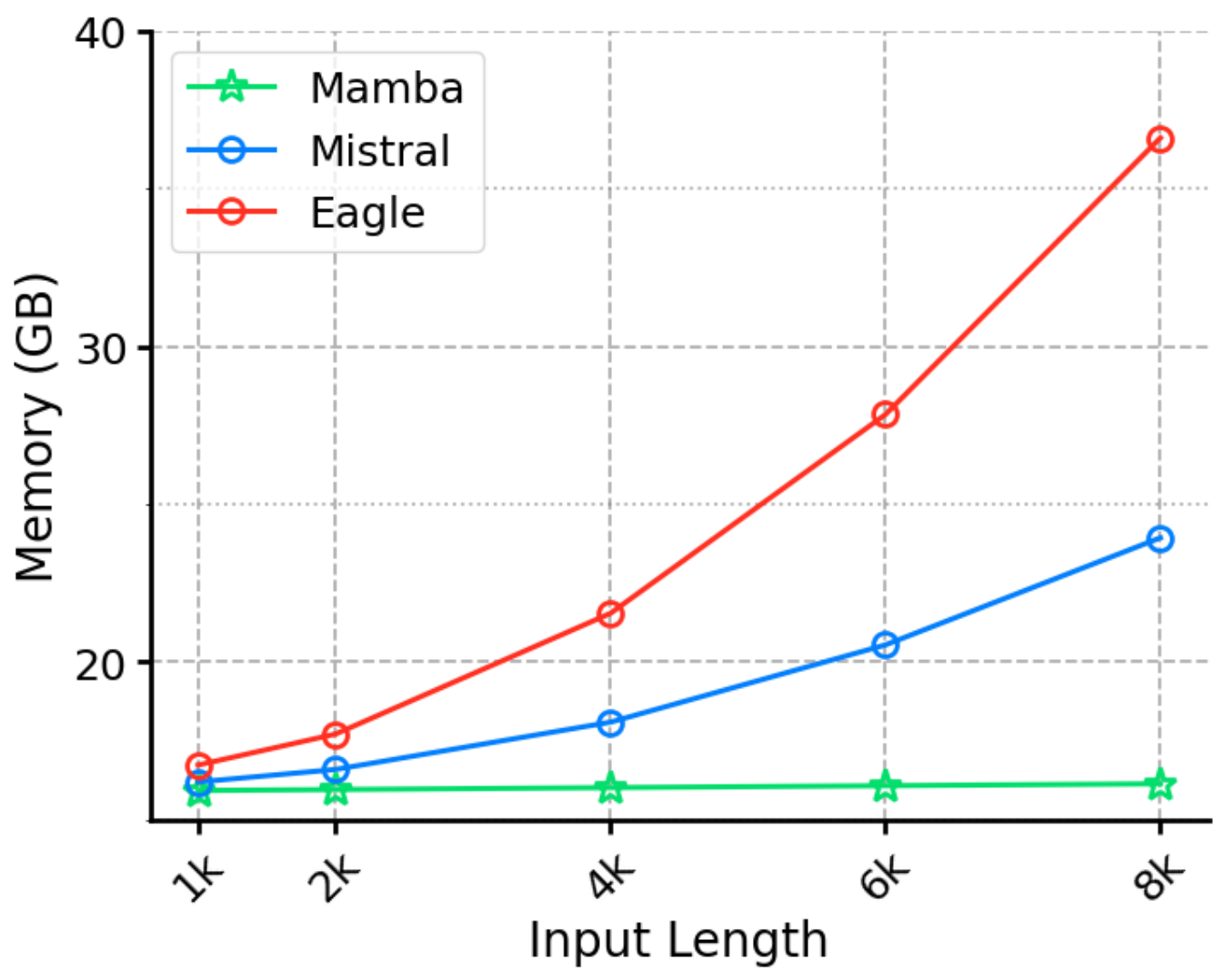}
        \centering
        \caption{Encoding Memory}\label{fig:encoding-memory}        
    \end{subfigure}
    \hfill
\begin{subfigure}{0.32\textwidth}
\includegraphics[width=\linewidth]{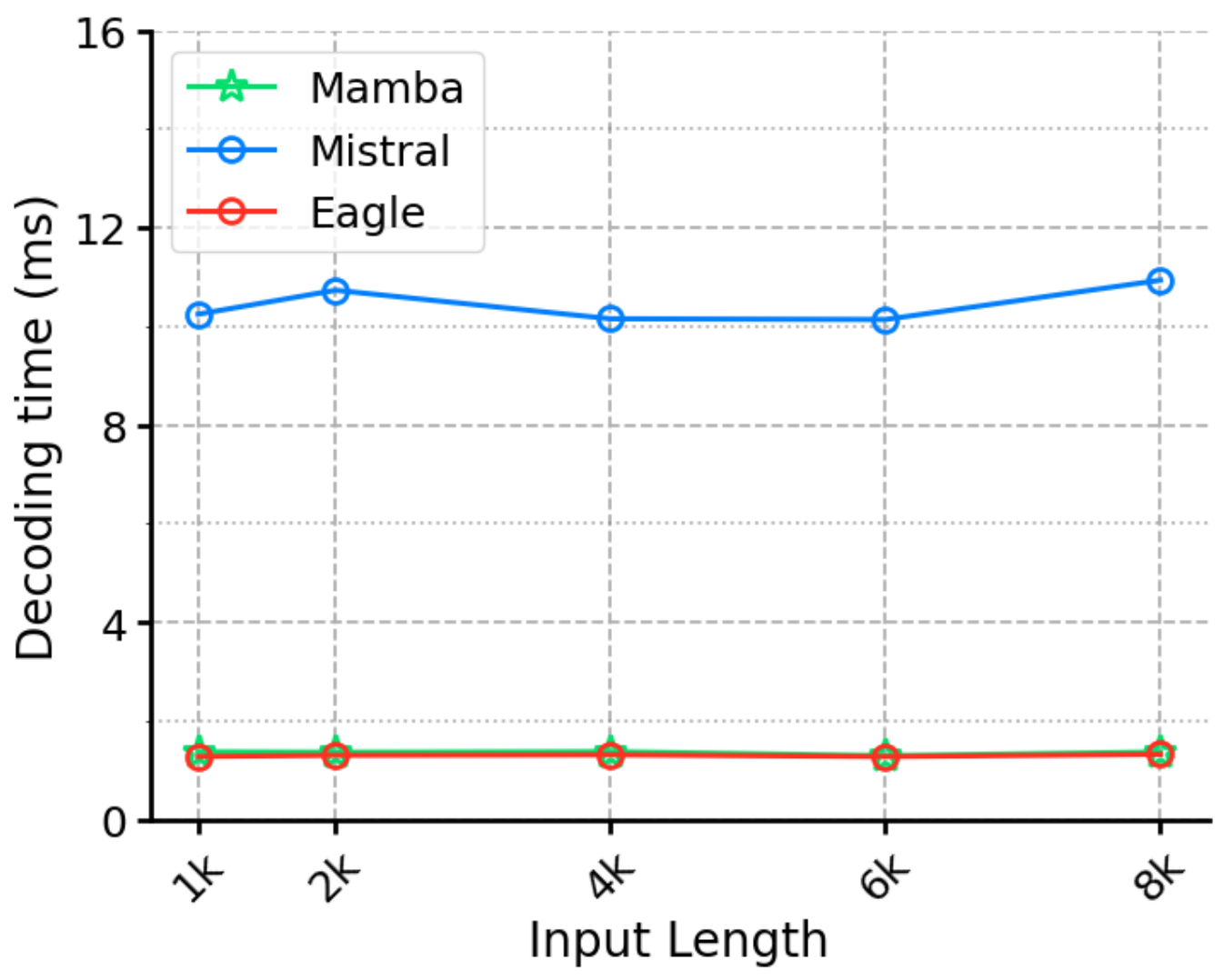} 
\caption{Decoding Time}\label{fig:decoding-time}
\end{subfigure}
    \begin{subfigure}{0.32\textwidth}
\includegraphics[width=\linewidth]{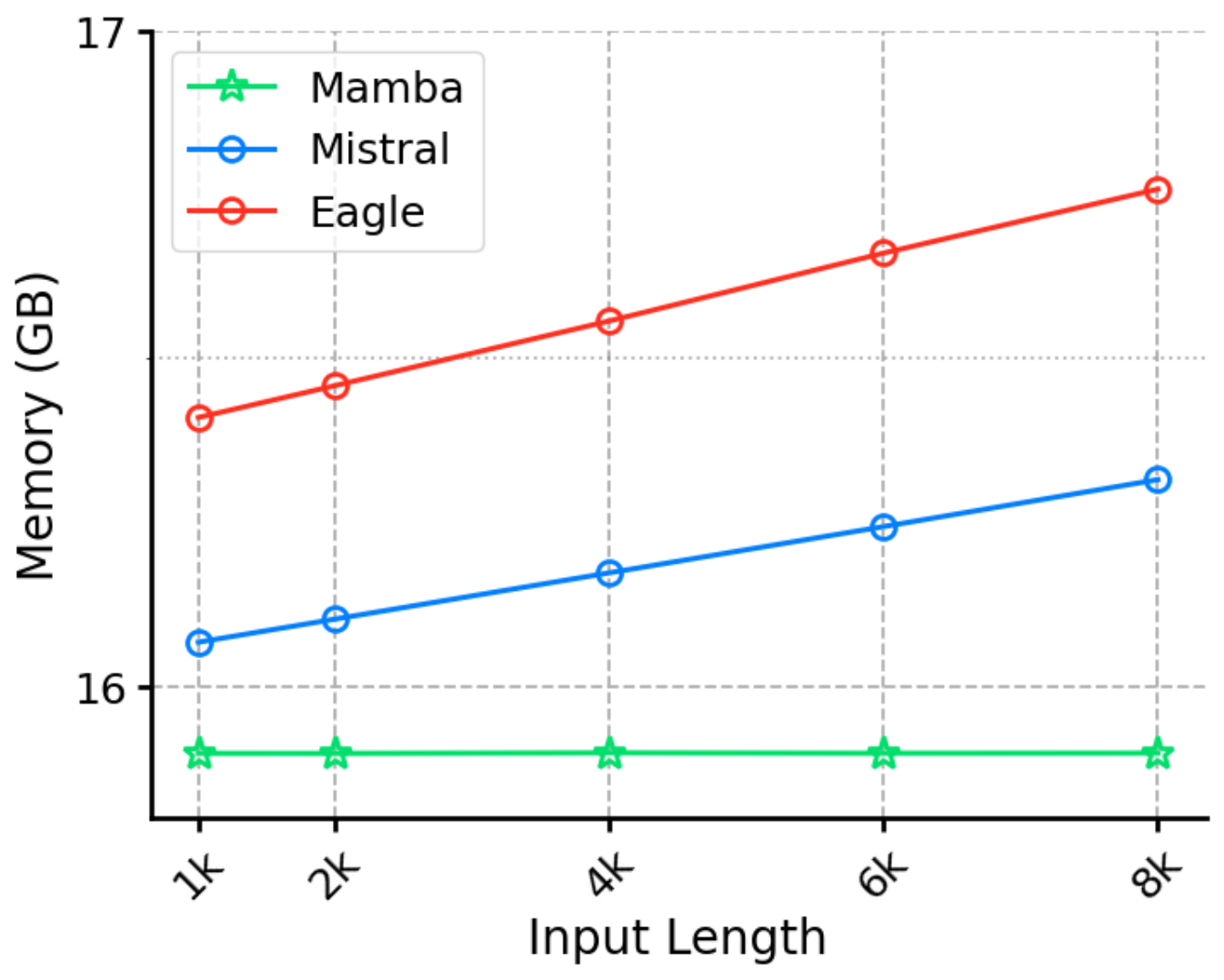}\caption{Decoding Memory}\label{fig:decoding-memory} 
\end{subfigure}
\caption{\textbf{Comparison of drafting time \& peak memory usage during encoding and decoding.} Mamba drafter maintains nearly constant decoding speed and memory usage, whereas both EAGLE, which employs self-speculation with a single-layer Transformer within the large target model, and the Mistral-based external drafter, which also utilizes a Transformer, exhibit substantially higher memory requirements as the context length increases. Here, the target model is Mistral-7B, and we consider a 160M-sized Mistral and a 130M-sized Mamba. Measurements are taken on an NVIDIA H100.}
    \label{fig:three_images}
    \hfill
    \vspace{-0.10in}
\end{figure*}

This raises a key question:
How can we develop an external drafter to 
have cross-model adaptability while also avoiding the limitation of the Transformer's quadratic computation for fast drafting?
This naturally leads us to explore non-quadratic sequential models such as 
state-space models (SSMs) \citep{gu2022efficiently}, as external drafters.
SSMs leverage a linear recurrence structure with a fixed state size, ensuring per-token computation and memory complexity remain constant during inference, making them more effective drafters than Transformers.
Specifically, we use Mamba~\citep{gu2023mamba}, a state-of-the-art SSM, as a drafter in SD and make the following key observations:
\noindent
\begin{itemize}
\item \textit{Mamba is an efficient external drafter, showing comparable results with self-speculation}: Mamba's drafting latency is comparable to self-speculation, with both latency and memory usage remaining low even for significantly longer input contexts, in contrast to alternative external drafters.
\item \textit{A smaller Mamba can often be more effective than a larger Transformer as an external drafter}: Despite its size, a small Mamba model achieves a comparable acceptance length to larger Transformers, with higher overall throughput due to its fast drafting.
\end{itemize}
To further leverage Mamba’s efficiency, we propose simple yet effective tree decoding strategies with Mamba drafters by formulating decoding as a multi-armed bandit (MAB)~\citep{slivkins2019mab} problem.
Specifically, we introduce a test-time tree search algorithm that dynamically optimizes the draft tree structure based on the input query. With Mamba's low latency, we observe that it benefits from trees of varying widths and lengths (see Table~\ref{tab:tree-config}). By framing the selection of the optimal tree structure as an MAB problem, we enable stable and adaptive adjustments of the drafting tree to accommodate different query types.

We conduct a comprehensive set of experiments to evaluate our Mamba-based approach, focusing on practical SD scenarios across a wide range of tasks~\cite{Narayan2018xsum,zheng2023mtbench,chen2021humaneval}. Our results demonstrate that Mamba-based drafting can significantly outperform traditional Transformer-based approaches. 
For example, our approach surpasses their throughput by 2x while having similar acceptance length.
Furthermore, our approach achieves throughput comparable to EAGLE~\citep{li2024eagle}, a recent single-layer Transformer drafter designed for a specific target model, in long-context scenarios while consuming up to 20 GB less memory.
This is notable as our target-agnostic Mamba drafter works with an arbitrary target model without re-training, whereas EAGLE, a self-speculation method, requires re-training of the drafter whenever the target model is updated.
Moreover, advances in SSMs, e.g., Mamba-2~\citep{dao2024mamba2}, directly benefit our approach, further enhancing the advantages of using an effective, target-agnostic drafter.

\section{Related Work}
\label{sec:related-works}

\paragraph{State-space models (SSMs).}
State-space models are strong linear models that combine the classical state-space representation~\citep{kalman1960new} with recurrent networks~\citep{elman1990finding}.
In contrast to models with quadratic scaling, such as Transformers~\citep{vaswani2017attention}, which use self-attention and experience increasing computational costs with sequence length, SSMs leverage linear recurrence~\citep{gu2022efficiently,gu2022parameterization,mehta2023long}, enabling more efficient training and inference.
This efficiency enables SSMs to excel, particularly in processing long sequences~\citep{tay2021long}.

Recent advancements, such as Mamba~\citep{gu2023mamba}, leverage hardware-aware algorithms and selection mechanisms to enhance SSMs further.
These developments have enabled SSMs to demonstrate effectiveness in complex tasks across diverse domains, including language, audio, and video~\citep{zhu2024vision,li2024videomamba,li2024mamba}. Building on this foundation, our research aims to utilize Mamba's efficiency for drafting, enabling faster and more effective speculative decoding.

\paragraph{Speculative decoding.}
Speculative decoding follows a draft-and-verify framework~\citep{stern2018blockwise}, where a smaller drafter generates candidate tokens that are verified by the target model. This method accelerates generation by increasing parallelism while ensuring alignment with the target model’s distribution. Later advancements extended this approach to sampling settings~\citep{leviathan2023fast,chen2023accelerating} and incorporated various optimization techniques to improve efficiency.

Speculative decoding approaches can be broadly categorized into two types. One approach utilizes external drafter models \citep{leviathan2023fast}, which provide high flexibility, allowing a single drafter to be directly used for different target models. On the other hand, self-speculation~\citep{cai2024medusa,li2024eagle} takes a different approach by utilizing a very small model that uses the target model's internal hidden states for drafting. While being faster, they require expensive re-training of the drafter for every target model, showing limited flexibility. In this work, we demonstrate that Mamba can get the best of both worlds, being an external drafter with very fast drafting speed.

To further improve the acceptance probability of the drafts, recent works move from sequential drafting to tree-structured drafting \citep{miao2024specinfer,yang2024multi, li2024eagle2}, which allows verifying multiple draft candidates in parallel. Additionally, researchers have explored non-Transformer drafters~\citep{he2023rest,fu2024break}. 
While \citet{wang2024mamba} introduced hardware optimizations for applying SSMs to speculative decoding, we explored the effectiveness on Transformer-based target models (which is a \emph{de facto} architecture) and developed an efficient inference scheme (i.e., tree-drafting) for Mamba. Furthermore, we present a more thorough comparison and analysis across different drafter models.

\section{Why Mamba for Speculative Decoding?}
\label{sec:observation}

\begin{figure}[t]
\centering\small
\centering
\includegraphics[width=0.95\linewidth,height=0.65\linewidth]{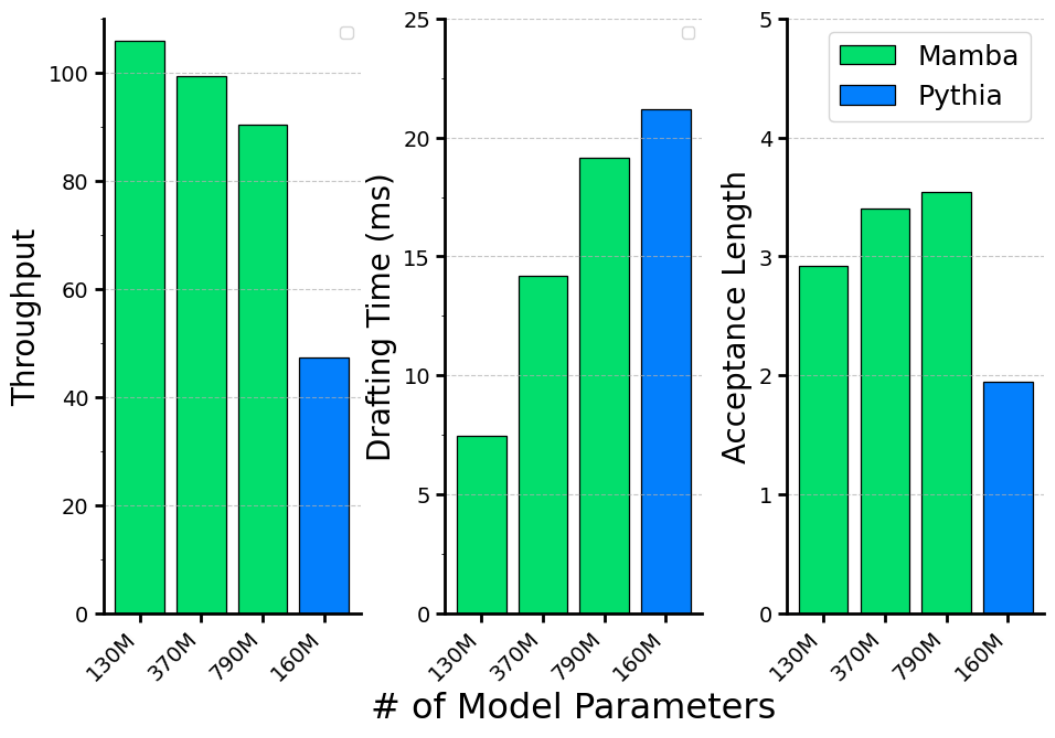} 
\vspace{-0.05in}
\caption{\textbf{Comparison of draft efficiency on GSM-8K.} Mamba drafters achieve substantially higher throughput than a Transformer drafter due to their faster drafting speed and favorable acceptance length. SD is run with a temperature of 1.0 and a draft length of 5.}
\label{fig:block-efficiency}
\end{figure}

In this section, we demonstrate that Mamba can serve as powerful external drafters for speculative decoding (SD).
We examine this in terms of both efficiency and effectiveness.
For \textit{efficiency}, we compare latency and peak memory usage during drafting, which includes encoding (prefill), initial forwarding of the given input sequence, and per-token generation during decoding (in \cref{fig:three_images}).
For \textit{effectiveness}, we report throughput, the average number of tokens per unit time, and acceptance length, the average number of tokens accepted per forward pass of the target model (in \cref{fig:block-efficiency,fig:alignment}).

\paragraph{Preliminaries: Speculative decoding.} Given a target model $M_p$, an efficient drafter $M_q$ and an input sequence $x_{\text{prefix}}$, SD starts by drafter $M_q$ generates candidate tokens with a length of $\gamma$ tokens, denoted as $\tilde{x}_1, \tilde{x}_2, ..., \tilde{x}_{\gamma}$, where each are sampled from drafter distribution, $\tilde{x}_i\sim q_i(x \mid \tilde{x}_1,..., \tilde{x}_{i-1}, x_{\text{prefix}})$. After drafting, candidate tokens along with $x_{\text{prefix}}$ are passed to the target model $M_q$ in parallel and corresponding target distribution $p_i(x \mid \tilde{x}_1,..., \tilde{x}_{i-1}, x_{\text{prefix}})$ and tail distribution $p_{\gamma+1}$ is obtained. Finally, each candidate token $\tilde{x}_i$ is verified by criterion sequentially from $i=1$ to $\gamma$. Here, the criterion is determined based on $p_i$ and $q_i$. Once a token is rejected at the position of $i$, a new token is sampled from the adjusted distribution $p^\prime_i$. If all candidates are accepted, an extra token is sampled from $p_{\gamma+1}$.

\subsection{Efficiency of Mamba as a drafter}
\label{sec:observation-fast}

To evaluate the efficiency of Mamba as the drafter, we compare it against two baselines: an external Transformer drafter and a self-speculation drafter. 
Specifically, we use Mistral-7B \citep{jiang2023mistral} as the target model and Mistral-160M, a smaller model with the same architecture, as the external Transformer drafter. For self-speculation, we employ EAGLE \citep{li2024eagle}, while Mamba-130M serves as the Mamba-based drafter. All drafters are trained from scratch, with additional training details provided in \Cref{app:train-details}. All models are instruction-tuned.

As \cref{fig:encoding-memory} illustrates, Mamba offers significant drafting efficiency compared to both baselines.
For example, as the input length increases, prefill memory for both Mistral and EAGLE grow nearly quadratically, while Mamba maintains memory usage with its efficient selectivity algorithm~\cite{gu2023mamba}.
Furthermore, as in \cref{fig:decoding-time}, Mamba exhibits significantly lower decoding latency than Mistral of similar size, and is even comparable to EAGLE.
Lastly, \cref{fig:decoding-memory} shows that Mamba's use of a single state enables it to maintain constant memory usage independent of the input length.
In contrast, other drafters require KV cache for decoding, causing the cache size to grow linearly with the input length, leading to high memory overhead.
These results demonstrate that Mamba drafters can make SD more adaptable to varying input lengths more effectively than Transformer drafters.

\subsection{Effectiveness of Mamba as a drafter}
\label{sec:observation-effective}

\begin{figure}[t]
\centering\small
\centering
\includegraphics[width=0.95\linewidth,height=0.622\linewidth]{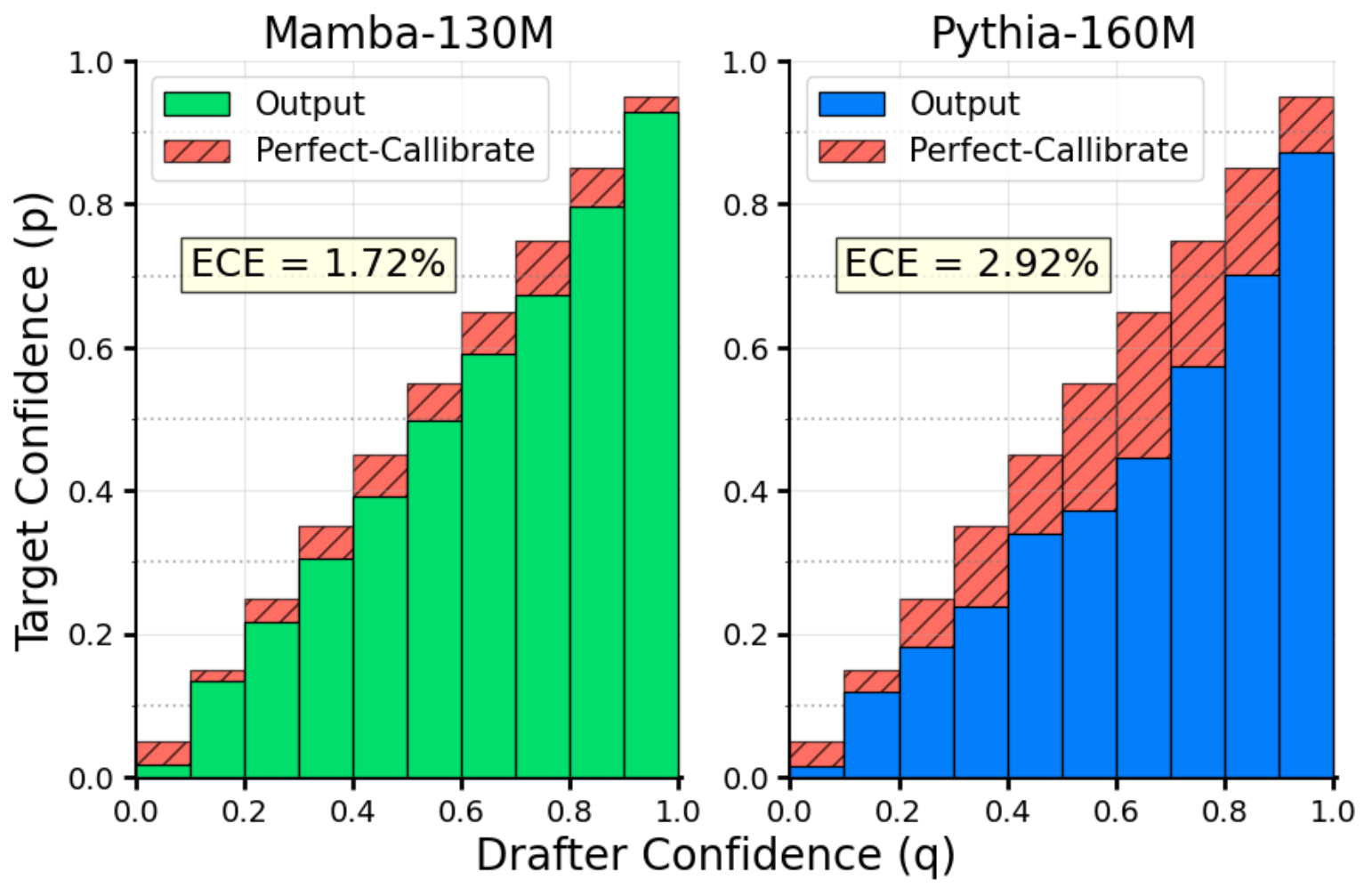} 
\vspace{-0.05in}
\caption{\textbf{Comparison of draft model calibration.} Reliability diagrams show that a small Mamba drafter aligns better with the target model, Pythia-6.9B, than the Transformer drafter, Pythia-160M, achieving a lower expected calibration error (ECE) on the XSum dataset.
}
\label{fig:alignment}
\end{figure}

In this section, we show that Mamba is an effective drafter for SD. 
For effective SD, there is an important trade-off between the drafter's speed and size, i.e., a larger drafter may achieve a higher acceptance length than smaller models by generating candidate tokens that are better aligned with the target model's distribution but increase the latency due to the larger size.
In this context, we observe that a small Mamba model can be a more effective drafter than a Transformer and, depending on the task, even larger Mamba models.
As shown in \cref{fig:block-efficiency}, the smallest Mamba achieves the highest throughput among all drafters due to its fast drafting and reasonable acceptance length.
Notably, the small Mamba achieves a higher acceptance length than the Transformer of similar size. This can be attributed to its better alignment with the distribution of the larger Transformer model, as illustrated in \cref{fig:alignment}.
Most interestingly, we found that smaller Mamba can be a stronger drafter than larger Mambas, as smaller Mamba shows significant drafting speed compared to larger sizes, generating more candidates with a slightly lower acceptance rate than larger models. We believe this highlights the exceptional drafting speed of Mamba, where such a phenomenon is not observed in Transformer-based drafters (see \cref{tab:main-pre}). 
To leverage this fast drafting speed further, we suggest a tree search algorithm to generate a better candidate and improve the acceptance length.

\section{Tree-Structured Drafting with Mamba}
\label{sec:method}

In this section, we introduce an effective drafting strategy for Mamba drafters
by using \emph{tree-structured decoding}, i.e., hierarchically expanding multiple candidate nodes at each step instead of sequentially generating tokens.
Specifically, we suggest an efficient way to implement tree search for Mamba decoding (in \cref{sec:method-tree}) and introduce a test-time tree searching algorithm to adaptively optimize draft tree structure (in \cref{sec:method-mab}).

\subsection{Tree-structured drafting with Mamba}
\label{sec:method-tree}
To improve the effectiveness of the drafter $M_q$, previous approaches \cite{yang2024multi, li2024eagle} sample multiple candidates from $q_i$ at each drafting step $i$ by constructing a draft tree. 
Especially in Transformer drafter, this process is accelerated by tree attention \citep{miao2024specinfer}, a specialized attention algorithm that represents the causal relationships between all tokens in the tree, thereby eliminating overlapped token forwarding, e.g., input sequence $x_{\text{prefix}}$. 
Here, we suggest an efficient tree-structured drafting specialized for Mamba.

\paragraph{Efficient tree-structured drafting with batch generation.} We demonstrate that Mamba can perform tree-structured drafting efficiently by using batch generation. Specifically, as Mamba only requires the current state to predict the next token (as it is a recurrent network), generating multiple nodes from the current node only requires copying the current state and then performing sampling. In contrast, Transformers are required to copy the current sequence length of the KV cache to predict the next token, and this overhead grows with the sequence length, making it crucial to eliminate such duplication using tree attention.

Formally, given a tree configuration $\mathcal{T} = (N_1, N_2, ..., N_{\gamma})$, where $\gamma$ is the draft length, and $N_i$ can be understood as the number of new nodes obtained by sampling from each node at the $i^{th}$ generation, one can view tree-structured drafting as a batch generation of a total batch size $\mathcal{B}_i=N_1 \times N_2 \times \cdots \times N_i$.

\paragraph{Efficient cache utilization for batch generation.} 
While efficient, batch generation indeed increases the computation complexity by a factor of $\mathcal{B}_i$ per generation, compared to sequential drafting, i.e., $N_1=...=N_{\gamma}=1$. 
To alleviate overheads from the batch size $\mathcal{B}$ during tree-structured drafting for Mamba, we propose a batch-wise cache implementation for Mamba. 
Specifically, given a tree configuration $\mathcal{T}$, we determine the possible batch sizes $\mathcal{B}_1, ..., \mathcal{B}_{\gamma}$ for each drafting positions. 
Using these calculated sizes, we create a state cache per each batch size and allocate memory in advance, preventing the memory re-allocation during duplication. Next, we leverage a graph cache \cite{Carilli2021cudagraphs} to accelerate the GPU computation flow for each batch size. This cache stores the graph structure of intermediate computations for each batch size, enabling efficient reuse of the computational graph across multiple executions. The reason this is feasible is that Mamba receives a fixed size of input $(\mathcal{B}_1, 1), (\mathcal{B}_2, 1), ..., (\mathcal{B}_\gamma, 1)$,  owing to its linear recurrence structure.

\subsection{Test-time dynamic tree search using multi-armed bandit}
\label{sec:method-mab}

We now present a way to systematically allocate the given budget to find the effective tree configuration $\mathcal{T}$,
based on the observation that Mamba benefits from different tree configurations across tasks (see \cref{sec:experiment-ablation} for details). 
To this end, we suggest to
formalize the tree configuration search problem as a multi-armed bandit (MAB) problem and dynamically optimize the tree configuration at inference time.

\paragraph{Decoding as multi-armed bandit.} Following a previous work \cite{kim2024mab}, we define each drafting and verification step as a round in the multi-armed bandit (MAB) framework. Specifically, in each round $t$, drafter $M_q$ follows an policy $\pi$ that choose $k^\text{th}$ tree configuration $\mathcal{T}^{(t)}_k$ from the pre-defined tree configuration set $\mathcal{S} = \{\mathcal{T}_1, \mathcal{T}_2, ..., \mathcal{T}_K\}$.
Then it performs $\gamma$ generations and obtains a reward $r^{(t)}$, e.g., the number of accepted tokens. The goal of the MAB problem is to design an optimal policy $\pi^*$ that maximizes the expected cumulative reward $\mathbb{E} \left[ \sum_{t=1}^{T} r^{(t)} \right]$ over a total of $T$ rounds, where $T$ is determined by the completion of generation for a given query.

\paragraph{Optimization.}
To balance exploration and stable convergence in MAB, we utilize the UCB algorithm~\cite{auer2002ucb} as our policy $\pi$. It chooses  tree configuration $\mathcal{T}_{k^*}^{(t)}$ at round $t$ as follows:
\begin{equation}
k^* = \operatorname*{arg\,max}_{k \in \{1,..,K\}}\hat{r}_k^{(t)} + \lambda_{\text{UCB}}  \sqrt{\frac{2 \ln t}{n_k^{(t)}}},
\end{equation}
where $\hat{r}_k^{(t)}$ is a cumulative reward mean, i.e., $\sum_ {t=1}^tr^{(t)}_k$ and $n_k^{(t)}$ is the count numbers of $k^\text{th}$ configuration is selected up to round $t$. For reward $r^{(t)}_k$, we define it as follows:
\begin{equation}\label{eq:reward}
r_k^{(t)} := -\left( \frac{1}{N_\text{accept}} + \lambda_{\tt \gamma} \frac{\gamma^{(\mathcal{T}_k)}}{N_\text{accept}} \right) \cdot I, 
\end{equation}
where $I$ is an indicator function, which is $1$ when the $k^\text{th}$ configuration is selected and $0$ otherwise, and $N_\text{accept}$ is the number of accepted tokens at round $t$. Especially, $\gamma^{(\mathcal{T}_k^{(t)})}$ represent draft length of selected tree $\mathcal{T}_k^{(t)}$ to penalize increase of draft times, as $\gamma$ is increase. We notice this reward directly originated from the SD speedup objective (see \cref{appendix:impl} for more details).

\begin{table*}[!ht]
\label{tab:main}
\centering
\small
\caption{
\textbf{Comparison with Transformer-based external drafters and self-speculation.} We evaluate SD using (a) pre-trained and (b) instruction-tuned models with both greedy decoding (temperature = 0) and sampling (temperature = 1). All drafters leverage tree-structured drafting and our method additionally uses the proposed tree search algorithm.
Throughput is reported along with the acceptance length shown in parentheses. The best results are shown in bold.
}
\vspace{-0.05in}
\begin{subtable}[t]{\textwidth} 
\centering\small
\caption{Pre-trained model}\label{tab:main-pre}
\resizebox{0.85\textwidth}{!}{\begin{tabular}{llccccccc}
\toprule
&\multicolumn{2}{c}{Drafter}&\multicolumn{3}{c}{Greedy (Temp=0)}&\multicolumn{3}{c}{Sampling (Temp=1)}\\
\cmidrule(lr){2-3}\cmidrule(lr){4-6} \cmidrule(lr){7-9}
Target & Method &Size & XSum & CNN-DM & GSM-8k & XSum & CNN-DM & GSM-8k\\
\midrule
\multirow{10}{*}{Pythia-6.9B} &
No drafter& $-$&53.30&49.29&54.69&52.51&45.33&53.81 \\
\cmidrule{2-9}
& \multirow{6}{*}{Pythia}&\multirow{2}{*}{70M}
&47.31&46.99&57.36&41.86&45.30&47.96\\&&& (1.52)&(1.54)&(1.68)&(1.67)&(1.76)&(1.77)\\
&&\multirow{2}{*}{160M}
& 50.05 & 49.53&67.89&46.67&47.17&55.40\\&& &(2.23)&(2.26)&(2.72)&(2.28)&(2.30)&(2.63)\\
& & \multirow{2}{*}{410M} 
&70.53&70.08&75.97&53.50&56.64&63.64\\&&& (4.62)&(4.73)&(4.64)&(3.60)&(3.80)&(4.01)\\
\cmidrule{2-9}
\rowcolor{blue!10} \cellcolor{white} 
& 
&&\textbf{138.80}&\textbf{131.97}&\textbf{149.46}&\textbf{108.68}&\textbf{105.01}&\textbf{119.67} \\
\rowcolor{blue!10} \cellcolor{white}
&
\multirow{-2}{*}{\cellcolor{blue!10}\textbf{Ours}}&\multirow{-2}{*}{\cellcolor{blue!10}130M}
&(4.55)&(4.38)&(4.57)&(3.53)&(3.53)&(3.73)\\
\midrule
\multirow{6.0}{*}{Mistral-7B} &
No drafter&$-$&51.15&49.55&50.31&53.49&47.40&52.92 \\
\cmidrule{2-9}
& \multirow{2}{*}{Mistral}& \multirow{2}{*}{160M}
& 61.55 & 61.04 & 49.38 &53.91&50.50&62.29\\
& &&(3.13)&(3.05)&(2.21)&(2.74)&(2.68)&(2.94)\\
\cmidrule{2-9}
\rowcolor{blue!10}
\cellcolor{white}
& &
&\textbf{76.71}&\textbf{65.23}&\textbf{77.50}&\textbf{79.18}&\textbf{70.95}&\textbf{82.63}\\
\rowcolor{blue!10}\cellcolor{white}
& \multirow{-2}{*}{\cellcolor{blue!10}\textbf{Ours}}&\multirow{-2}{*}{\cellcolor{blue!10}130M}
 &(2.39)&(2.13)&(2.25)&(2.73)&(2.65)&(2.73)\\

\bottomrule
\end{tabular}
}
\end{subtable}
\begin{subtable}[t]{\textwidth} 
\centering\small
\vspace{0.15in}
\caption{Instruction-tuned model}\label{tab:main-inst}
\resizebox{0.9\textwidth}{!}{\begin{tabular}{llc cccccc}
\toprule
&\multicolumn{2}{c}{Drafter}&\multicolumn{3}{c}{Greedy (Temp=0)}&\multicolumn{3}{c}{Sampling (Temp=1)}\\
\cmidrule(lr){2-3}
\cmidrule(lr){4-6} \cmidrule(lr){7-9}
Target & Method & External? & MT-bench & Alpaca & Human-Eval & MT-bench & Alpaca & Human-Eval\\
\midrule
\multirow{8.5}{*}{Pythia-6.9B} &
No drafter&$-$&54.51&55.28&54.76&53.89&54.72&54.21 \\
\cmidrule{2-9}
& \multirow{2}{*}{Pythia}&\multirow{2}{*}{\cmark}
&70.71& 60.77& 109.51&65.73&62.07&109.52\\
&  &&(3.10)&(2.65)&(4.68)&(3.03)&(2.82)&(4.25)\\
\cmidrule{2-9}
& \multirow{2}{*}{EAGLE}
&\multirow{2}{*}{\xmark}&\underline{125.61}&\textbf{117.17}&\underline{122.44}&\underline{87.01}&\underline{78.58}&\underline{83.05}\\
&&&(3.85)&(3.53)&(4.71)&(2.67)&(2.40)&(2.97)\\
\cmidrule{2-9}
\rowcolor{blue!10}
\cellcolor{white}
&
& 
&\textbf{128.21}&\underline{114.08}&\textbf{172.38}&\textbf{110.20}&\textbf{108.54}&\textbf{143.55}\\
\rowcolor{blue!10}\cellcolor{white}
& \multirow{-2}{*}{\cellcolor{blue!10}\textbf{Ours}} & \multirow{-2}{*}{\cmark}
&(3.91)&(3.41)&(5.41)&(3.65)&(3.51)&(4.82)\\
\midrule
\multirow{8.5}{*}{Mistral-7B} &
No drafter&$-$&52.97&53.58&52.30&52.39&53.02&52.34 \\
\cmidrule{2-9}
& \multirow{2}{*}{Mistral}
&\multirow{2}{*}{\cmark}&67.47&61.40&100.23&57.19&51.05&80.94\\
&&&(3.04)&(2.73)&(4.53)&(2.84)&(2.40)&(3.92)\\
\cmidrule{2-9}
& \multirow{2}{*}{EAGLE}
&\multirow{2}{*}{\xmark}&\textbf{107.16}&\underline{94.03}&\textbf{132.69}&\textbf{94.03}&\textbf{86.60}&\textbf{122.11}\\
&&&(3.22)&(2.79)&(3.98)&(2.90)&(2.63)&(3.78)\\
\cmidrule{2-9}
\rowcolor{blue!10}
\cellcolor{white}
& &
&\underline{102.48}&\textbf{96.83}&\underline{118.04}&\underline{88.68}&\underline{82.75}&\underline{87.81}\\
\rowcolor{blue!10}\cellcolor{white}
& \multirow{-2}{*}{\cellcolor{blue!10}\textbf{Ours}}& \multirow{-2}{*}{\cmark} 
 &(3.16)&(2.96)&(3.69)&(2.95)&(2.71)&(2.94)\\

\bottomrule
\end{tabular}
}
\end{subtable}
\end{table*}

\section{Experiments}
\label{sec:experiments}

In this section, we present a comprehensive evaluation of our proposed Mamba drafter framework. 

\paragraph{Baselines.}
We use a wide range of Transformer-based drafters as baselines, applying tree drafting for a fair comparison with our Mamba drafter.
Additionally, we consider EAGLE~\citep{li2024eagle}, a recent single-layer Transformer drafter that leverages tree drafting and is directly trained to align with the target model, as a baseline for self-speculation methods.

\paragraph{Evaluation metrics.}
To compare the gains in decoding acceleration from SD across different drafter types, we focus on throughput, which is the number of tokens generated per second during inference, as a measure of overall inference speed. For a more comprehensive evaluation of effectiveness, we also report the average acceptance length, which indicates the average number of tokens accepted per forward pass of the target model.


\begin{table*}
\centering
\small
\caption{\textbf{Comparisons on LongBench.} Throughput (tokens/s) is the primary metric, with acceptance length shown in parentheses.
We also report peak memory, calculated by summing the memory consumption of both the target and drafter models during the prefill phase.
Bold indicates the best result, while the runner-up is underlined.
}\label{table:long-context}
\vspace{-0.05in}
\begin{subtable}[t]{\textwidth} 
\centering\small
\begin{tabular}{lccccccccc|cccc}
\toprule
&&\multicolumn{4}{c}{Single-Document QA}&\multicolumn{4}{c}{Multi-Document QA}&\multicolumn{4}{c}{Peak Memory (GB)}\\
\cmidrule(lr){3-6} \cmidrule(lr){7-10} \cmidrule(lr){11-14}
Method&External?&1k&2k&4k&8k&1k&2k&4k&8k&1k&2k&4k&8k\\
\midrule 
No drafter&-&31.02&27.89&24.35&19.30&28.17&24.22&19.01&14.83&15&16&20&36\\
\midrule
\multirow{2}{*}{Mistral}&\multirow{2}{*}{\cmark}&25.30&23.28&19.48&15.23&24.64&21.06&16.49&12.18&\multirow{2}{*}{31}&\multirow{2}{*}{33}&\multirow{2}{*}{38}&\multirow{2}{*}{59}\\
&&(2.43)&(2.37)&(2.24)&(2.21)&(2.53)&(2.48)&(2.44)&(2.39)&\\
\midrule
\multirow{2}{*}{EAGLE}&\multirow{2}{*}{\xmark}&
\underline{53.13}&\textbf{47.00}&\textbf{37.27}&\textbf{26.12}&\underline{42.48}&\underline{35.38}&\underline{25.10}&\underline{17.36}&\multirow{2}{*}{32}&\multirow{2}{*}{34}&\multirow{2}{*}{42}&\multirow{2}{*}{72}\\
&&(2.73)&(2.81)&(2.76)&(2.71)&(2.60)&(2.61)&(2.68)&(2.64)
\\
\midrule
\rowcolor{blue!10}&&\textbf{55.09}&\underline{45.65}&\underline{36.32}&\underline{24.92}&\textbf{47.91}&\textbf{36.40}&\textbf{26.27}&\textbf{17.77}&&&& \\
\rowcolor{blue!10}
\multirow{-2}{*}{\cellcolor{blue!10}\textbf{Ours}}&\multirow{-2}{*}{\cellcolor{blue!10}\textbf{\cmark}}&(2.91)&(2.77)&(2.77)&(2.80)&(2.94)&(2.86)&(2.88)&(2.87)&\multirow{-2}{*}{31}&\multirow{-2}{*}{32}&\multirow{-2}{*}{36}&\multirow{-2}{*}{52}
\\
\bottomrule
\end{tabular}
\end{subtable}
\end{table*}

\subsection{Comparison with transformer drafters}
First, we compare our Mamba drafter with Transformer external drafters.

\paragraph{Pre-trained models.}
We evaluate the performance of the pre-trained Mamba drafter across various language modeling tasks. Specifically, we consider XSum~\cite{Narayan2018xsum} and CNN-DailyMail~\cite{hermann2015cnndm} for general language modeling tasks, as well as GSM-8K~\cite{cobbe2021gsm8k} for mathematical language modeling. 
As summarized in \cref{tab:main-pre}, while larger drafters like Pythia-410M achieve slightly better throughput gains on some datasets due to increased acceptance length, small Transformer drafters (e.g., Pythia-70M) show minimal improvement over the vanilla autoregressive baseline without SD, with only marginal benefits in the sampling case.
In contrast, Mamba significantly improves throughput across datasets and temperature settings.
For instance, in GSM-8K, Mamba achieves nearly 2x the throughput of Pythia-410M on sampling setup.
Notably, even when the Mamba drafter has a lower acceptance length than Transformer drafters, it still outperforms them due to its fast drafting speed.

\paragraph{Instruction-tuned models.}
Next, we evaluate the Mamba drafter in instruction-following scenarios using MT-bench~\citep{zheng2023mtbench} for multi-turn dialogues, Alpaca~\cite{taori2023alpaca} for general instruction-following tasks, and HumanEval~\citep{chen2021humaneval} for code generation.
As shown in \cref{tab:main-inst}, the Mamba drafter outperforms Transformer drafters across all instruction-following datasets.
For example, in HumanEval, while Pythia and Mistral drafters improve throughput over the vanilla autoregressive baseline by 54.75 and 47.93 for their respective target models, Mamba achieves even more significant gains of 117.62 and 65.74 for these models. These results highlight Mamba's flexibility and superior generalization to diverse instruction-following tasks compared to Transformer drafters.

\subsection{Comparison with self-speculation}
\label{sec:exp-recent}
We demonstrate that the Mamba drafter, which is an external drafter, can achieve competitive throughput even against recent approaches that train drafters with direct access to target models.
Specifically, we consider EAGLE, which uses a single-layer Transformer drafter trained to generate tokens from the target model’s last hidden states for better alignment with the target model.
\cref{tab:main-inst} reports the results on instruction-following tasks, where our Mamba drafter achieves throughput gains comparable to EAGLE across the datasets and target models.
Notably, on MT-bench, the Mamba drafter achieves a throughput of 125.61, which is comparable to EAGLE's 128.21 for Pythia-6.9B.
These results highlight not only Mamba's fast drafting speed (see \cref{fig:decoding-time}), but also its effectiveness in achieving comparable acceptance length without requiring access to target models.

\subsection{Long-context scenarios}
\label{sec:exp-long}
To evaluate Mamba's scalability in long-context scenarios, we conduct SD experiments on LongBench~\citep{bai2023longbench} using input lengths ranging from 1k to 8k, with all drafters trained with the same context limit.\footnote{Following EAGLE, we fine-tune pre-trained Mamba and Mistral on ShareGPT with a context limit of 2k.}
Additionally, we apply 
YaRN~\citep{peng2023yarn} to extend the context limit for
both Transformer drafters and EAGLE.
As shown in \cref{table:long-context}, which presents the results on LongBench, Mamba maintains a higher acceptance length on longer inputs compared to both Transformer drafters and EAGLE, even when the latter utilizes YaRN to extend the context length.
In the Single-Document QA task, as the input length increases from 1k to 8k, Mistral's acceptance length decreases from 2.43 to 2.21, while Mamba remains more stable, changing from 2.91 to 2.80.
This stability reflects Mamba's ability to extrapolate effectively via recurrence~\citep{gu2023mamba}.
Moreover, Mamba generalizes well to unseen complex distributions.
In the Multi-Document QA task (which is a complex problem compared to single-document QA as the answer is located across multiple documents), Mamba consistently achieves throughput gains comparable to EAGLE.
Here, we notice that the gains are obtainable more efficiently: when applying to 8k, Mamba only consumes memory up to 52GB, compared to EAGLE, which reaches up to 72GB (with both including the memory needed for drafter and target verification).


\begin{table}[t]
\centering
\caption{
\textbf{Cross-target model performance on MT-bench.}
Experiments are run with a temperature of 0 and sequential drafting with a length of 5.
}
\label{tab:cross-model}
\vspace{-0.05in}
\small
\resizebox{0.485\textwidth}{!}{
\begin{tabular}{cccc}
\toprule
Method & Setup & Accept length &Throughput\\
\midrule
\multirow{2}{*}{EAGLE}&Pythia $\rightarrow$ Pythia&2.59&94.75\\
&Mistral $\rightarrow$ Pythia&N/A&N/A\\
\midrule
\rowcolor{blue!10}&Pythia $\rightarrow$ Pythia&3.08&112.69\\
\rowcolor{blue!10}\multirow{-2}{*}{\cellcolor{blue!10}\textbf{Ours}}&Mistral $\rightarrow$ Pythia&2.45&93.20\\
\bottomrule
\end{tabular}
}
\end{table}


\begin{table}[t]
\centering
\caption{\textbf{Tree-structured drafting on MT-bench.} Experiments are run with a temperature of 0 and a fixed tree configuration of $(3,2,2,1,1)$. Tree drafting yields notable improvements in all performance metrics.}
\label{tab:comp-multi}
\vspace{-0.05in}
\small
\begin{tabular}{cccc}
\toprule
Tree? & Accept length & Latency & Throughput\\
\midrule
\xmark&3.08&6.62&112.69\\
\rowcolor{blue!10}
\cmark&3.91&8.30&127.37\\
\bottomrule
\end{tabular}
\end{table}

\begin{table}[t]
\centering
\caption{\textbf{Throughput by tree configuration.} Throughput (tokens/s) for different tree configurations given as  $(N_1, ..., N_\gamma)$, where $N_i$ indicates the number of samples at $i^\text{th}$ generation in drafting step.}
\label{tab:tree-config}
\vspace{-0.05in}
\small
\begin{tabular}{cccc}
\toprule
Method&(3,3,2,1)&(3,2,2,1,1)&(2,2,2,1,1,1)\\
\midrule
Pythia&75.38&70.71&
63.75\\
\rowcolor{blue!10}\textbf{Ours}&124.99&
127.37&
124.37\\
\bottomrule
\end{tabular}
\end{table}

\begin{table}[t]
\centering
\caption{\textbf{Effects of test-time tree search on throughput.} Experiments are run with a temperature of 0.}
\label{tab:mab}
\vspace{-0.05in}
\small
\resizebox{0.47\textwidth}{!}{
\begin{tabular}{cccc|c}
\toprule
Search? & MT-bench & Alpaca & HumanEval & Avg.\\
\midrule
{\xmark}&124.99&\textbf{116.12}&149.15&130.09\\
{\cmark}&\textbf{128.21}&114.08&\textbf{172.38}&\textbf{138.22}\\
\bottomrule
\end{tabular}}
\end{table}

\subsection{Cross-target model performance}
Using an external drafter enables plug-and-play integration with new target models without the need to re-train the drafter for each specific model.
To evaluate Mamba's performance as an external drafter, we use the Mamba drafter with a target model that the drafter has not been explicitly trained to align with.
Specifically, we use the instruction-tuned variant of Pythia-6.9B as the target model and Mamba trained with the Mistral-7B tokenizer.
As shown in \cref{tab:cross-model}, even without explicit training, Mamba achieves throughput comparable to EAGLE, which is specifically trained for the target model.
This highlights Mamba’s flexibility as an external drafter, enabling efficient deployment without the need for costly re-training.

\subsection{Ablations and analysis}
\label{sec:experiment-ablation}
We further evaluate the contributions of individual components in our framework to the gains in decoding acceleration. Here we mainly consider throughput (tokens/sec) as the performance metric. 

\paragraph{Tree-structured drafting.}
In \cref{tab:comp-multi}, we show the impact of tree-structured drafting on performance. 
Our approach improves acceptance length with minimal drafting latency overhead, resulting in higher throughput gains. This effect is similar to the tree-attention mechanism used in Transformer-based drafters. 
These results demonstrate that our batch generation enables Mamba to benefit from tree-structured drafting, which has not been explored in the field to the best of our knowledge.

\paragraph{Tree configurations.}
\cref{tab:tree-config} shows that Mamba maintains stable throughput across diverse configurations, whereas Transformer drafters exhibit a decline as tree length increases. This is due to Mamba's very fast drafting, which effectively mitigates the overhead of using longer trees.

\paragraph{Test-time tree search.} We further analyze the impact of the test-time tree search algorithm. Specifically, we use as tree candidates $(3,3,2,1)$, $(3,2,2,1,1)$, and $(2,2,2,1,1,1)$ and compare them with naive tree-structured drafting that utilizes a fixed tree configuration, i.e., $(3,2,2,1)$. As shown in \cref{tab:mab}, our multi-armed bandit (MAB)-based algorithm often improves throughput significantly on several datasets (e.g., HumanEval) compared to the naive approach.
\section{Conclusion}
\label{sec:conclusion}

In this work, we present Mamba-based drafters as an effective solution to the challenges of existing speculative decoding methods. Leveraging the linear structure of state-space models, Mamba significantly improves drafting speed and memory efficiency. To further enhance drafting quality, we introduce a novel tree-based search algorithm. 
Our experimental results show that Mamba-based drafters not only outperform existing external drafting techniques but also match the performance of advanced self-speculation approaches, particularly in long-context scenarios.

\section*{Limitations}

While our Mamba-based drafting approach demonstrates significant improvements in inference speed, there are several opportunities for further enhancement. Although Mamba-based drafters require memory for hidden state maintenance, this overhead could potentially be optimized through efficient memory management techniques. While our tree-based search method's performance depends on hyperparameter settings, this flexibility allows for customization across different use cases, and future work could develop adaptive optimization strategies. Additionally, while beyond the scope of this work, investigating self-drafting capabilities within a single Mamba model—leveraging its efficient architecture to serve as both drafter and verifier—represents an intriguing direction for future research. These limitations and opportunities point to exciting directions that could further advance the efficiency of large language model inference.

\bibliography{custom}

\newpage
\appendix
\onecolumn

\section{Experimental Details}
\subsection{Datasets}
We evaluate six benchmarks: three for pre-trained models and three for instruction-tuned models. These include XSum \cite{Narayan2018xsum} and CNN-DailyMail \cite{hermann2015cnndm} for general language modeling, GSM-8K \cite{cobbe2021gsm8k} for mathematical reasoning, MT-Bench \cite{zheng2023mtbench} for multi-turn dialogues, Alpaca \cite{taori2023alpaca} for general instruction-following, and HumanEval \cite{chen2021humaneval} for code generation. Following EAGLE \cite{li2024eagle}, we subsample each dataset to approximately 80 samples.

For evaluating longer-context scenarios, we use six tasks from LongBench \cite{bai2023longbench} for document-based question answering: three (NarrativeQA, Qasper, MultifieldQA-en) for Single-Document QA and three (HotpotQA, 2WikiMultihopQA, MuSiQue) for Multi-Document QA. While LongBench originally includes Chinese-language tasks (MultifieldQA-zh, DuReader), we observe that the target model produced poor outputs on these tasks and therefore exclude them. For pre-processing, we first filter out data samples with input lengths exceeding 8k tokens and truncate them to specific input lengths, such as 1k, 2k, 4k, and 8k.

\subsection{Architectures}
\paragraph{Pre-train models.} For the pre-trained target model, we consider EleutherAI/pythia-6.9b and mistralai/Mistral-7B-v0.1.
For external Transformer drafters, we use smaller models from the same family as the target model, specifically Pythia-70M, 160M, 410M and Mistral-160M.
For Mamba drafters, we use two versions of Mamba-130M: Mamba-Pythia-130M and Mamba-Mistral-130M, which share tokenizers with Pythia and Mistral, respectively.
In cases where no official pre-trained checkpoints are available, e.g., Mistral-160M and Mamba-Mistral, we pre-train them from scratch (see \cref{app:train-details} for details).

\paragraph{Instruction-tuned models.}
For the instruction-tuned target model, we consider allenai/open-instruct-pythia-6.9b-tulu and mistralai/Mistral-7B-Instruct-v0.1, which are instruction-tuned from EleutherAI/Pythia-6.9b and mistralai/Mistral-7B-v0.1, respectively. To obtain instruction-tuned drafters, we supervised fine-tune (SFT) the pre-trained drafters on ShareGPT, following the training dataset used for EAGLE (see \cref{app:train-details} for details). Additionally, we obtain corresponding EAGLE for Pythia and Mistral  by following their official released training code.

\subsection{Training Details.}\label{app:train-details}
Following a common pre-training recipe \footnote{\url{https://github.com/facebookresearch/lingua}}, we pre-train Mistral-160M and Mamba-Mistral-130M on FineWeb-Edu \cite{penedo2024fineweb} for 5,000 training steps using a batch size of 4,096 and a context limit of 2k. Next, we fine-tune (SFT) the pre-trained drafters on ShareGPT for 2 epochs with a batch size of 128 and a context limit of 2k, following the standard SFT procedure \footnote{\url{https://github.com/huggingface/alignment-handbook/blob/main/recipes/zephyr-7b-beta/README.md}}. For validation, we use HellaSwag \cite{zellers2019hellaswag}, ARC-Easy \cite{allenai2018arc}, and PIQA \cite{bisk2020piqa}. We select the best model by testing various learning rates, specifically \{2e-3, 2e-4, 2e-5\}.

\subsection{Implementation}
\label{appendix:impl}
\paragraph{Tree-structured drafting.} 
Following previous work~\cite{yang2024multi}, we implement tree-structured drafting for the external Transformer drafter. We use a tree configuration with a depth of 5, i.e., (3,2,2,1,1), to align the draft length with EAGLE. For EAGLE, we directly follow its official tree-structured drafting implementation.

\paragraph{Reward modeling for MAB.}
To obtain the reward function in \cref{eq:reward}, we directly use the speed up formula from SD per drafting step. Given a target model and drafter's decoding time, i.e., per-token generation time, as $T_\text{ target}$ and $T_\text{draft}$, the total time of SD $T_\text{total}^\text{SD}$ per drafting step is as follows:
\begin{equation}
T_\text{total}^\text{SD} = T_\text{target}(\gamma) + \gamma \cdot T_\text{draft},
\end{equation}
where $\gamma$ is draft length, and $T_\text{target}(\gamma)$ is verification time for forwarding $\gamma$ draft tokens. Then, we compute SD's  decoding time $T_\text{Avg}^\text{SD}$ by dividing the number of accepted tokens $N_\text{accept}$, i.e., $T_\text{Avg}^\text{SD} = \frac{T_\text{Total}^\text{SD}}{N_\text{accept}}$. Finally, the speed up of SD per drafting step is as follows:
\begin{equation}
\text{speedup} = \frac{T_\text{target}}{T_\text{Avg}^\text{SD}}
= N_\text{accept}\cdot \frac{T_\text{target}}{T_\text{target}(\gamma) + \gamma \cdot T_\text{draft}}
\end{equation}
Then, the inverse of speedup is as follows:
\begin{equation}
\frac{1}{\text{speedup}} = \frac{1}{N_\text{accept}} \cdot\frac{T_\text{target}(\gamma)}{T_\text{target}} + 
\frac{\gamma}{N_\text{accept}}\cdot\frac{T_\text{draft}}{T_\text{target}}
\end{equation}
Generally, we can assume $\frac{T_\text{target}(\gamma)}{T_\text{target}} \simeq 1$, as $\gamma$ is not larger value, and $\frac{T_\text{draft}}{T_\text{target}} \simeq \lambda_{\gamma}$ as it is constant during drafting. Then, our reward function $r$ is derived as follows:
\begin{equation}
r = \frac{1}{N_\text{accept}}  + 
\lambda_{\gamma} \cdot \frac{\gamma}{N_\text{accept}}
\end{equation}
This formula originated from the inverse of speed up, so we need to minimize this function.

\subsection{Greedy Decoding and Sampling}
Algorithm \ref{verification_algorithm} outlines the verification process for draft token acceptance using two decoding strategies:

\begin{enumerate}
    \item \textbf{Greedy decoding} (red, lines 7, 15, 22): This method selects tokens deterministically by setting the temperature to zero, effectively forcing the model to choose the most probable token at each step. This is equivalent to using the one-hot version of the target model, $p_\text{one-hot}$.
    \item \textbf{Sampling-based approach} (blue, lines 8, 16, 23): In contrast, this method introduces stochasticity by sampling tokens from the probability distribution given by the target model $p$. This allows for more diverse outputs.
\end{enumerate}

The verification algorithm works by comparing the probabilities assigned by the target model and the draft model. The acceptance of a draft token $\tilde{x}_t$ depends on the ratio of these probabilities. If the draft token is rejected, a new token is sampled based on either the greedy or sampling-based approach.

\begin{algorithm}
\caption{Verification Algorithm.}
\label{verification_algorithm}
\begin{algorithmic}[1]
\State \textbf{Given} target model $p$, one-hot version of target model $p_\text{one-hot}$, and draft model $q$.
\State \textbf{Given} input sequence $x_\text{prefix}$, and draft sequence $\tilde{x}$ of length $\gamma$.
\For{$t = 1$ to $\gamma$}
    \State Sample $u$ from a uniform distribution:
    \State \hspace{2em} $u \sim U[0,1]$

    \State Get the probability of each model for the draft token $\tilde{x}_t$
    \State  {\color{red} \hspace{2em} $p_t = p_\text{one-hot}(\tilde{x}_t|x_\text{prefix}, x_{1}, \dots, x_{t-1})$}
    \State  {\color{blue} \hspace{2em} $p_t = p(\tilde{x}_t|x_\text{prefix}, x_{1}, \dots, x_{t-1})$}
    \State \hspace{2em} $q_t = q(\tilde{x}_t|x_\text{prefix}, x_{1}, \dots, x_{t-1})$
    
    \If{$u < \min\left(1, \frac{p_t}{q_t}\right)$}
        \State Accept the token $\tilde{x}_t$:
        \State \hspace{2em}  $x_{t} \gets \tilde{x}_t$.
    \Else
        \State Reject the draft token $\tilde{x}_t$ and sample a new one:
        \State {\color{red} \hspace{2em}  $x_{t} \sim p_\text{one-hot}(x|x_\text{prefix}, x_{1},  \dots, x_{t-1})$.}
        \State {\color{blue} \hspace{2em}  $x_{t} \sim \left(p(x|x_\text{prefix}, x_{1},  \dots, x_{t-1}) - q(x|x_\text{prefix}, x_{1},  \dots, x_{t-1})\right)_\text{+}$.}
        \State \textbf{break}
    \EndIf
\EndFor
\If{all tokens are accepted}
    \State Sample an extra token $x_{\gamma+1}$:
    \State {\color{red} \hspace{2em} $x_{\gamma+1}\sim p_\text{one-hot}(x|x_\text{prefix}, x_{1}, \dots, x_{\gamma})$.}
    \State {\color{blue} \hspace{2em} $x_{\gamma+1}\sim p(x|x_\text{prefix}, x_{1}, \dots, x_{\gamma})$.}
\EndIf
\State \textbf{Output} the accepted token sequence $x_{1}, \dots, x_{n}$, where $n$ is the accepted token length.
\end{algorithmic}
\end{algorithm}

\subsection{Computational Resources}
We conduct most experiments on a single NVIDIA RTX 4090 24GB GPU, except for longer-context experiments in \cref{table:long-context}, where we use a single NVIDIA H100 80GB GPU to efficiently handle input lengths of up to 8k tokens. For pre-training, we leverage 8 NVIDIA H200 141GB GPUs, which takes approximately one day. For instruction-tuning of external drafters and training EAGLE, we use 8 NVIDIA RTX 4090 24GB GPUs, requiring approximately two hours and one day, respectively. Here, we remark that training EAGLE incurs additional computational cost, as it requires extracting hidden states from the training data via forward passes through the target model.

\end{document}